\def\year{2020}\relax
\documentclass[letterpaper]{article} % DO NOT CHANGE THIS
\usepackage{aaai20}  % DO NOT CHANGE THIS
\usepackage{times}  % DO NOT CHANGE THIS
\usepackage{helvet} % DO NOT CHANGE THIS
\usepackage{courier}  % DO NOT CHANGE THIS
\usepackage[hyphens]{url}  % DO NOT CHANGE THIS
\usepackage{graphicx} % DO NOT CHANGE THIS
\usepackage{graphicx}  % DO NOT CHANGE THIS
\usepackage{pdfpages}

\title{Weakly Supervised Disentanglement by Pairwise Similarities}
\author{ \Large \textbf{Junxiang Chen,} 
\Large \textbf{Kayhan Batmanghelich} 
\\ 
Department of Biomedical Informatics\\
University of Pittsburgh, Pittsburgh, PA 15232, US\\ %If you have multiple authors and multiple affiliations
\{juc91,kayhan\}@pitt.edu % email address must be in roman text type, not monospace or sans serif
}

\newcommand{\citet}[1]{\citeauthor{#1} \shortcite{#1}}
\newcommand{\citep}{\cite}

\usepackage[utf8]{inputenc} % allow utf-8 input
\usepackage{booktabs}       % professional-quality tables
\usepackage{amsfonts}       % blackboard math symbols
\usepackage{nicefrac}       % compact symbols for 1/2, etc.
\usepackage{microtype}      % microtypography

\usepackage[utf8]{inputenc} % allow utf-8 input
\usepackage{booktabs}       % professional-quality tables

\usepackage{tikz}
\usetikzlibrary{fit,positioning}

\usepackage{filecontents}
\usepackage{amsmath}
\usepackage{array}

\usepackage{graphicx}
\usepackage{subcaption}

\usepackage{capt-of}

\usepackage{tikz}
\usetikzlibrary{fit,positioning}

\usepackage{multirow}
\usepackage{mathrsfs}

\usepackage{multirow}

\usepackage{enumitem}
\setlist{nolistsep}

\usepackage{adjustbox}

\usepackage{algorithm}
\usepackage{algorithmic}

\newcommand\ie {{\it i.e., }}

\usepackage{dsfont}

\makeatletter
\def\maketag@@@#1{\hbox{\m@th\normalfont\normalsize#1}}
\makeatother

\begin{document}

\maketitle
\let\thefootnote\relax\footnote{ The code is available at \url{https://github.com/batmanlab/VAE_pairwise}. }

\begin{abstract}
Recently, researches related to unsupervised disentanglement learning with deep generative models have gained substantial popularity. However, without introducing supervision, there is no guarantee that the factors of interest can be successfully recovered~\cite{locatello2018challenging}. Motivated by a real-world problem, we propose a setting where the user introduces weak supervision by providing similarities between instances based on a factor to be disentangled. The similarity is provided as either a binary (yes/no) or a real-valued label describing whether a pair of instances are similar or not. We propose a new method for weakly supervised disentanglement of latent variables within the framework of Variational Autoencoder. Experimental results demonstrate that utilizing weak supervision improves the performance of the disentanglement method substantially.
\end{abstract}

\section{Introduction}
Disentanglement learning is a task of finding latent representations that separate the explanatory factors of variations in the data \cite{bengio2013representation}.
In recent years, several methods \cite{higgins2017beta,kim2018disentangling,chen2018isolating,lopez2018information} have been proposed to solve disentanglement learning under the Variational Autoencoder (VAE) framework. However, most of these existing methods are unsupervised. In this paper, we focus on improving the disentangling performance by utilizing weak supervisions in terms of pairwise similarities.  

\citet{locatello2018challenging} showed that unsupervised disentanglement learning is fundamentally impossible if no inductive biases on models and datasets are provided. Existing unsupervised methods control the implicit inductive biases by choosing the hyperparameters. However, the factor of interest is not guaranteed to be successfully recovered by only tuning the hyper-parameters. Providing strong supervisions with discrete or real-valued labels have been previously suggested~\cite{narayanaswamy2017learning,kulkarni2015deep}. However, such supervision can be expensive to acquire.

Our method is motivated by a real-world problem. In this problem, we want to understand how the Computer Tomography (CT) images are related to the severity of Chronic Obstructive Pulmonary Disease (COPD), which is a devastating disease related to cigarettes smoking. Since COPD manifests itself as airflow limitation, its severity can be measured via spirometry (meaning the measuring of breath). However, the disease severity is usually measured by combining two \cite{vestbo2013global} or three \cite{quanjer2012multi} spirometric measures. It is not obvious how we can represent disease severity with one real value. Therefore, we represent disease severity using real-valued pairwise similarities between subjects, which are computed based on spirometric measures. The available CT images and the pairwise similarities motivate us to develop a disentanglement method that utilizes pairwise similarities when analyzing images.

In this paper, we assume that we are provided a measure of similarity between instances based on a specific factor of interest, in addition to the observations. The pairwise similarity can be binary (yes/no) or real-valued and may only be provided for a few pairs of instances. The goal is to learn disentangled representations such that a subset of the latent variables explain the factor of interest, but do not convey information about other factors of variations. We propose to achieve this goal by constructing a VAE model that generates both the samples and the pairwise similarities based on latent representations. We achieve disentanglement by letting the pairwise similarities depend on a subset of the latent variables but independent of the other latent variables, and penalizing the information capacity of the dependent latent variables. Our empirical evaluations on several benchmark datasets and the COPD dataset show that providing pairwise similarities improves the performance of the disentanglement method substantially.

\noindent{\bf Contributions} \hspace{1mm}
We make the following contributions in this paper:
(1)~We design a latent variable model that enables a user to provide similarities between instances in the desired latent space.
(2)~The similarity can be a binary or real-valued value provided for all or a subset of the pairs of instances. We formulate the model with a VAE framework and propose an efficient algorithm to train the model.
(3)~We conduct extensive experiments on benchmark datasets and a real-world dataset. Experimental results demonstrate that introducing weak supervision improves the disentanglement performance in different tasks. 

\section{Background}
\label{sec:background}
\noindent{\bf $\boldsymbol{\beta}$-VAE} \hspace{1mm}
 The $\beta$-VAE \cite{higgins2017beta} is base for many disentanglement methods. It introduces the inductive bias by increasing the weight of the KL divergence term in the evidence lower bound (ELBO) objective function, defined as
\begin{equation}
    \label{equ:beta-VAE}
    \begin{aligned}
    \mathcal{L}_{\beta-VAE} = & \max_{\boldsymbol{\theta}, \boldsymbol{\phi}}
    \hspace{2mm}
    \mathbb{E}_{ \mathbf{x}_n \sim p_{\text{data}} } 
        [\mathbb{E}_{q_{\boldsymbol{\phi}}(\mathbf{z}_n|\mathbf{x}_n) }[\log p_{\boldsymbol{\theta}}(\mathbf{x}_n|\mathbf{z}_n)]
        \\
        & - \beta \mathcal{D}_{KL} 
        \left( q_{\boldsymbol{\phi}}(\mathbf{z}_n|\mathbf{x}_n) || p(\mathbf{z})
        \right)
        ]
        ,
    \end{aligned}
\end{equation}
where $\mathbf{X} = \left\{ \mathbf{x}_n \right\}_{n =1}^{N}$ and $ \mathbf{Z} = \left\{ \mathbf{z}_n \right\}_{n =1}^{N}$ denote the observed samples and  the corresponding latent variables respectively, and $N$ is the number of samples. We use $ p_{\boldsymbol{\theta}}(\mathbf{x}_n|\mathbf{z}_n)$ and $q_{\boldsymbol{\phi}}(\mathbf{z}_n|\mathbf{x}_n)$ to represent the decoder and encoder networks that are parameterized by $\boldsymbol{\theta}$ and $\boldsymbol{\phi}$, respectively. We let $\mathcal{D}_{KL}(\cdot || \cdot)$ denote the KL divergence and $p(\mathbf{z})$ denote prior distribution for $\mathbf{z}$. In this paper, we let $p(\mathbf{z})$ be an isotropic unit Gaussian distribution. In the equation, $\beta \ge 1$ is a hyperparameter that controls the weight for the KL divergence term.

\section{Method}
\label{sec:method}

%The purpose of this paper is to learn the disentangled representation by analyzing both the instances and the pairwise similarities. 
We assume that we have access to the noisy observations of the similarities for pairs of instances.
We use $ \mathbf{Y} = \{ y_{ij} \}_{ (i,j) \in \mathcal{J} }$ to represent the set of observed similarities,  where  $\mathcal{J} \subseteq \{ (i,j)| i, j \in \{ 1, \ldots ,N \} \}$. 
Note that not all pairwise similarity labels are necessarily observed. 
We allow $y_{ij}$ to be either binary ($y_{ij} \in \{ 0,1\}$) or real-valued between $0$ and $1$, where a larger value of $y_{ij}$ indicates a stronger similarity between $\mathbf{x}_{i}$ and $\mathbf{x}_{j}$.
%% ****** might be a sentence missing

In the following sections, we first explain the general framework of our model. We then discuss how the pairwise similarities can be incorporated into the model. Finally, we introduce a regularization term that encourages disentanglement.

\subsection{The General Framework}
\label{sec:setup}

\begin{figure}[t]
    \centering
    \includegraphics[width=.8 \columnwidth]{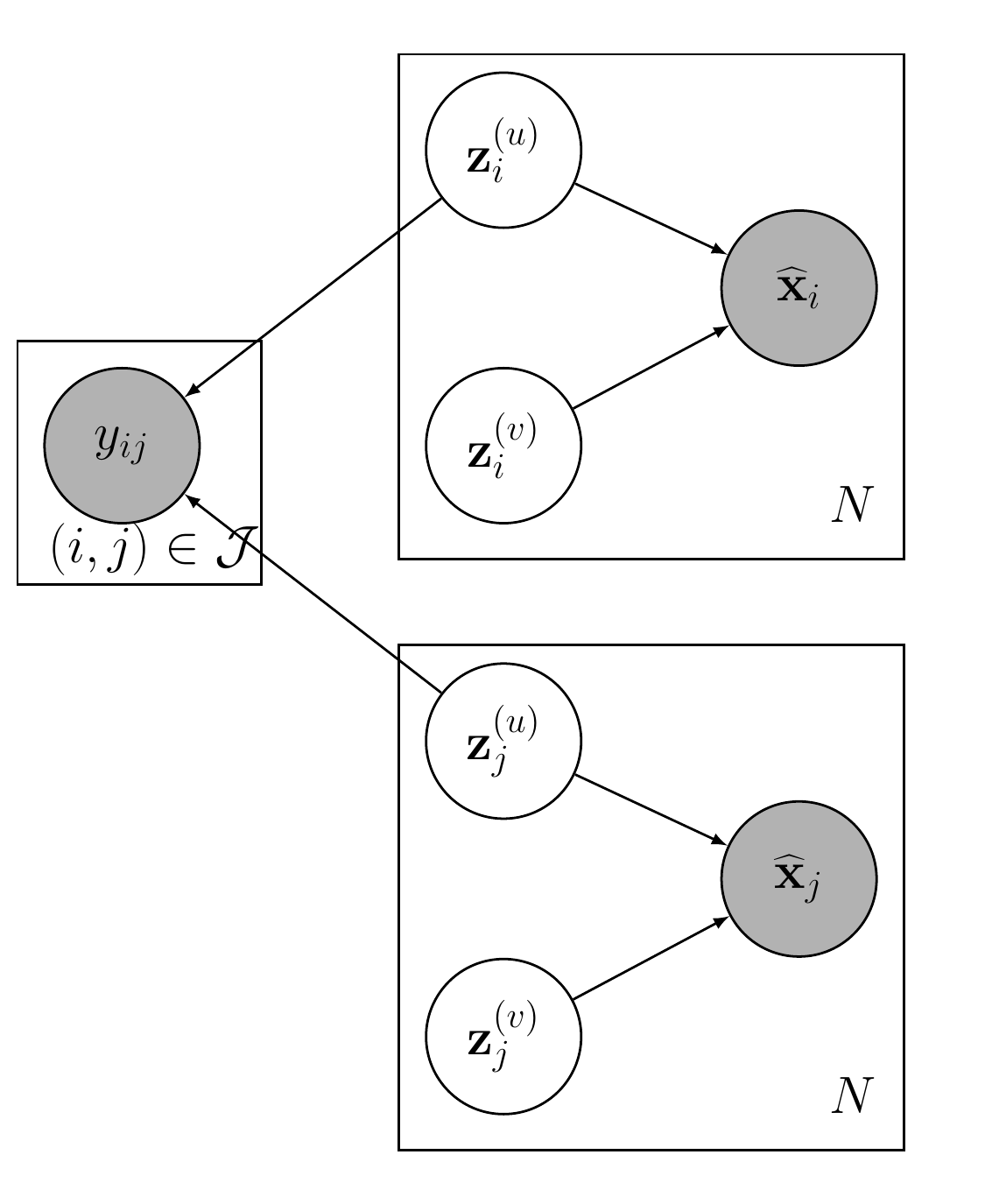}
    \caption{ The decoder model $p(\mathbf{X}, \mathbf{Y} | \mathbf{Z})$.}
    \label{fig:decoder}
    \includegraphics[width=.52 \columnwidth]{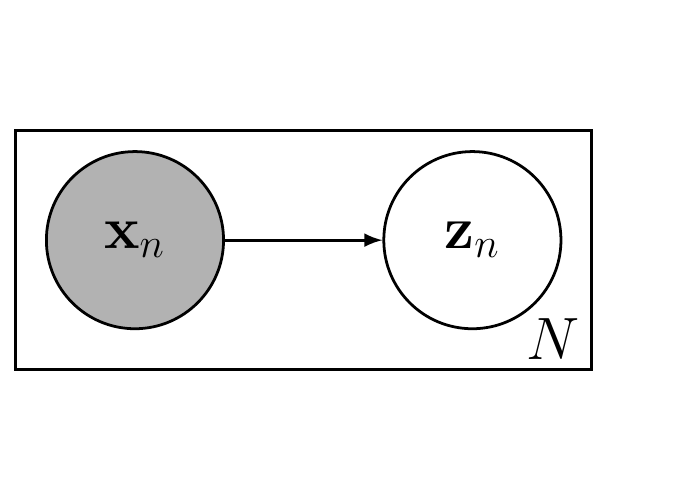}
    \caption{ The encoder model $q(\mathbf{Z}| \mathbf{X})$.}
    \label{fig:encoder}
\end{figure}

We assume that both $\mathbf{X}$ and $\mathbf{Y}$ are noisy observations; hence, we use a probabilistic approach to model uncertainty. 
We adopt the VAE framework \cite{kingma2013auto} such that $\mathbf{x}_n$ is reconstructed based on the latent variables $\mathbf{z}_{n}$.
We assume that the latent variable $\mathbf{z}$ is divided into two sub-spaces, {\it i.e., } $\mathbf{z} = [ \mathbf{z}^{(u)}, \mathbf{z}^{(v)} ] $, where  $\mathbf{z}^{(u)}$ (with $d^{(u)}$ dimensions) accounts for the latent variables \emph{relevant} to the factors of interest, while $\mathbf{z}^{(v)}$ (with $d^{(v)}$ dimensions) accounts for the rest of information.  
Since $y_{ij}$ represents pairwise similarity based on the factors of interest, it is only dependent on the coordinates of the latent variables of $\mathbf{x}_{i}$ and $\mathbf{x}_{j}$ in the $\mathbf{z}^{(u)}$ subspace; {\it i.e., } $p(y_{ij} | \mathbf{z}_i , \mathbf{z}_j) = p(y_{ij} | \mathbf{z}_i^{(u)} , \mathbf{z}_j^{(u)})$. Therefore, the joint distribution of the observed instances and similarities has the following factorization,
\begin{equation}
    \label{equ:decoder}
\begin{adjustbox}{width=.95\columnwidth,center}
{$
    p_{\boldsymbol{\theta}}(\mathbf{X}, \mathbf{Y}| \mathbf{Z}) = \prod_{n=1}^N p_{\boldsymbol{\theta}} ( \mathbf{x}_n | \mathbf{z}_n)
    \prod_{(i,j)\in \mathcal{J}} p(y_{ij} | \mathbf{z}^{(u)}_i, \mathbf{z}^{(u)}_j).
$}
\end{adjustbox}
\end{equation}
This model can be represented using a graphical model as shown in Figure \ref{fig:decoder}. In this equation,  $p_{\boldsymbol{\theta}}(\mathbf{x}_n | \mathbf{z}_n)$ represents the reconstruction model of the VAE framework. We explain $p(y_{ij} | \mathbf{z}_i^{(u)} , \mathbf{z}_j^{(u)})$ in the next section.

\subsection{Modeling Pairwise Similarity}
\label{sec:similarity}
We view $y_{ij}$ as the noisy observation of the similarity between $i$'th and $j$'th instances, which can be either a binary or a real-value measurement.  We use the following function to model conditional of $y_{ij}$ for both cases,
\begin{equation}
\begin{adjustbox}{width=\columnwidth,center}
{$
    \label{equ:y_real}
    p \left( y_{ij}| \mathbf{z}^{(u)}_i, \mathbf{z}^{(u)}_j \right) 
    = \frac{1}{\mathcal{C}} \left( g( \mathbf{z}^{(u)}_i, \mathbf{z}^{(u)}_j ) \right)^{y_{ij}}
    \left( 1 - g( \mathbf{z}^{(u)}_i, \mathbf{z}^{(u)}_j ) \right)^{1 - y_{ij}},
$}
\end{adjustbox}
\end{equation}
where $\mathcal{C}$ is the normalization constant and $g(\cdot,\cdot)$ is a function encoding the strength of the similarity given the relevant latent variables $\mathbf{z}_{i}^{(u)}$ and $\mathbf{z}_{j}^{(u)}$. In Equation (\ref{equ:y_real}), when $y_{ij}$ is a binary variable, $g$ can be viewed as probability that a user labels $y_{ij}$ as $1$. Hence, we choose $g$ to return a value between 0 and 1 and $\mathcal{C} = 1$. When $y_{ij}$ is real-valued between $0$~and~$1$, Equation (\ref{equ:y_real}) enables us to compute the normalization constant in a closed form:
\begin{equation}
    \begin{adjustbox}{width=.8\columnwidth,center}
    {$
        \begin{aligned}
            \mathcal{C} & = \int_0^1
            \left(  g( \mathbf{z}^{(u)}_i, \mathbf{z}^{(u)}_j ) \right)^{y_{ij}}
            \left( 1 - g( \mathbf{z}^{(u)}_i, \mathbf{z}^{(u)}_j ) \right) ^ {1 - y_{ij}}
            d y_{ij}
            \\
             & = \frac{2g( \mathbf{z}^{(u)}_i, \mathbf{z}^{(u)}_j ) - 1}
            { \log \left( g( \mathbf{z}^{(u)}_i, \mathbf{z}^{(u)}_j ) \right) 
              - \log \left( 1 - g( \mathbf{z}^{(u)}_i, \mathbf{z}^{(u)}_j ) \right)
            }.
        \end{aligned}
    $}
    \end{adjustbox}
\end{equation}

\begin{figure}[t]
    \centering
    \includegraphics[width=.8\columnwidth]{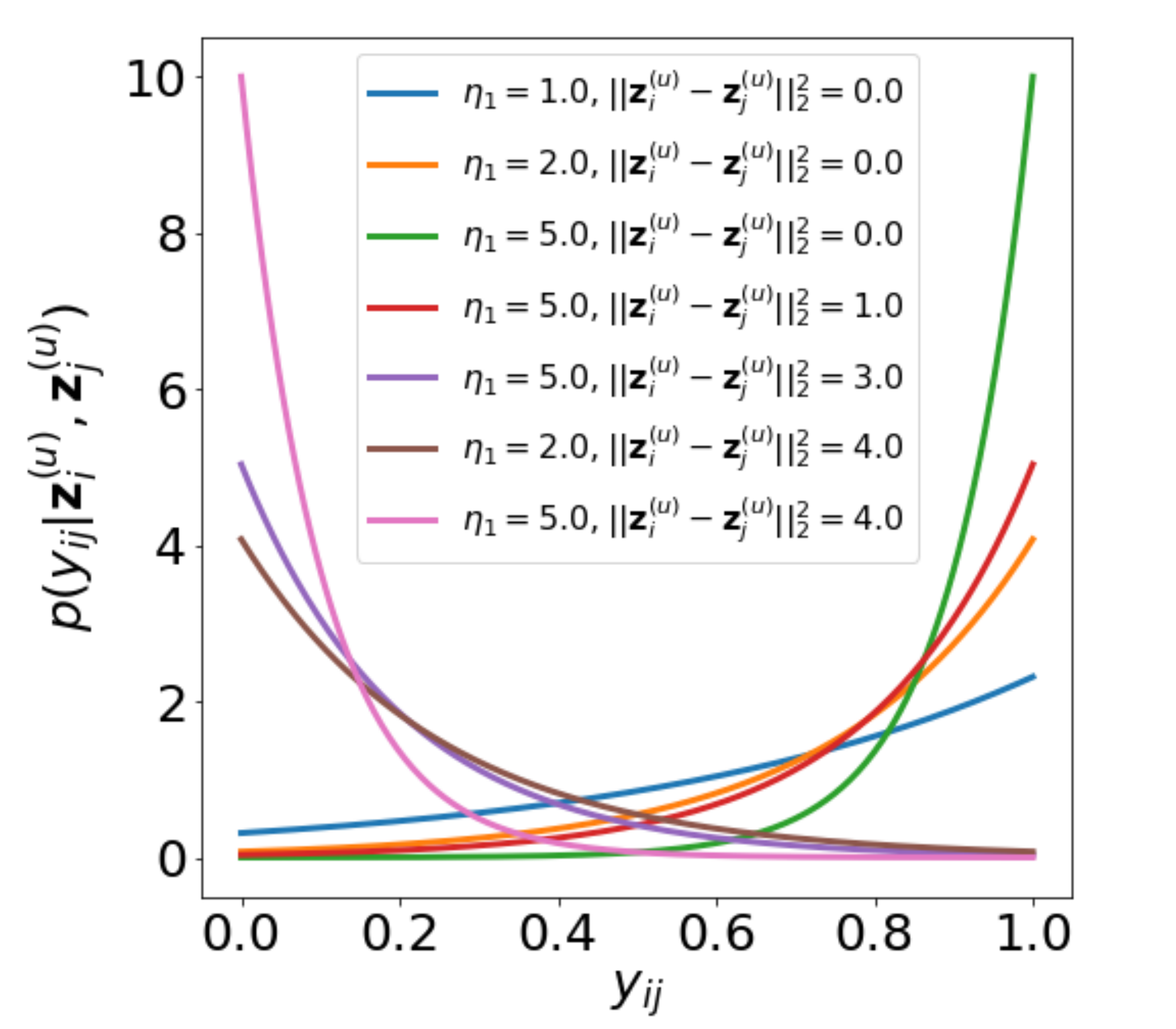}
    \caption{ Plot for $p ( y_{ij} | \mathbf{z}_i^{(u)}, \mathbf{z}_j^{(u)} ) $ for real-valued $y_{ij}$. We fix the thresholding hyperparameter $\eta_2 = 2$. When $||\mathbf{z}_i^{(u)} - \mathbf{z}_j^{(u)}||_2^2$ is small, it is more likely to have a large $y_{ij}$ and vice versa. The hyperparameter $\eta_1$ controls the ``steepness'' of the distribution. }
    \label{fig:sigmoid}    
\end{figure}

We adopt the following form for $g$:
\begin{equation}
    \label{equ:g}
    \begin{adjustbox}{width=.8\columnwidth,center}
    {$
    g \left( \mathbf{z}^{(u)}_i, \mathbf{z}^{(u)}_j \right) = \sigma \left( \eta_1 \left( \eta_2 -  ||\mathbf{z}_i^{(u)} - \mathbf{z}_j^{(u)}||_2^2 \right)  \right),
    $}
    \end{adjustbox}
 \end{equation}
 where $\eta_1$ and $\eta_2$ are positive real hyperparameters controling the ``steepness'' and ``threshold'' of the similarity, respectively; and $\sigma(\cdot)$ is the sigmoid function. When $\eta_1 \rightarrow \infty$,  $ g( \mathbf{z}^{(u)}_i, \mathbf{z}^{(u)}_j )$ can be regarded as a hard thresholding function indicating whether  or not $||\mathbf{z}_i^{(u)} - \mathbf{z}_j^{(u)}||_2^2$ is smaller than $\eta_2$. We replace the hard thresholding function with a sigmoid function $\sigma(\cdot)$ to make sure this function differentiable. The Figure~\ref{fig:sigmoid} shows that when $||\mathbf{z}_i^{(u)} - \mathbf{z}_j^{(u)}||_2^2$ is small, it is more likely to have a large $y_{ij}$ and vice versa.

 \subsection{Disentanglement via Regularization}
\label{sec:disentanglement}
Our goal of disentanglement is to encode all information about the factor of interest into $\mathbf{z}^{(u)}$ and to prevent it from containing irrelevant information.  The general idea is to limit the capacity of $\mathbf{z}^{(u)}$; hence, its capacity can be used only for the relevant factors. Similar to the $\beta$-VAE, we use a regularized ELBO that increases the weight of the KL divergence between the approximate posterior ({\it i.e., } $q_{\boldsymbol{\phi}}(\mathbf{z}^{(u)}_n | \mathbf{x}_n)$) and the prior ({\it i.e., } $p(\mathbf{z}^{(u)})$), but we do not impose extra regularization for $\mathbf{z}^{(v)}$. The regularization term is defined as
\begin{equation}
    \label{equ:regularization}
\begin{adjustbox}{width=.75\columnwidth,center}
{$
    \begin{aligned}
    \mathcal{R} = & \ - \mathbb{E}_{ \mathbf{x}_n \sim p_{\text{data}} } 
        \left[ 
            \beta \mathcal{D}_{KL} 
                \left( q_{\phi}( \mathbf{z} ^{(u)}_n |\mathbf{x}_n) || p(\mathbf{z}^{(u)} ) \right)
        \right]
        \\
          &\  -\mathbb{E}_{ \mathbf{x}_n \sim p_{\text{data}} } 
        \left[
            \mathcal{D}_{KL} 
                \left( q_{\phi}(\mathbf{z} ^{(v)}_n|\mathbf{x}_n) || p( \mathbf{z}^{(v)} ) \right)
        \right], 
    \end{aligned}
$}
\end{adjustbox}
    \end{equation}
where $\beta \ge 1 $ is a real-valued hyperparameter that controls the weight of KL divergence.

\subsection{Overall Model}

The overall objective can be written as follows, 
\begin{equation}
    \begin{adjustbox}{width=\columnwidth,center}
    {$
        \label{equ:objective}
        \begin{aligned}
        \mathcal{L} 
        = & \max_{\boldsymbol{\theta}, \boldsymbol{\phi}} \hspace{2mm}
            \mathbb{E}_{ \mathbf{x}_n \sim p_{\text{data}} } 
            \left[ \mathbb{E}_{q_{\boldsymbol{\phi}}(\mathbf{z}_n|\mathbf{x}_n) }[\log p_{\boldsymbol{\theta}}(\mathbf{x}_n|\mathbf{z}_n)]
            \right]
        \\
        & + \mathbb{E}_{ (i, j) \in \mathcal{J}} 
                        \left[ \mathbb{E}_{q_\phi(\mathbf{z}_i, \mathbf{z}_j |\mathbf{x}_i, \mathbf{x}_j) }
                            \left[
                                \log p \left( y_{ij}|\mathbf{z}_i^{(u)}, \mathbf{z}_j^{(u)} \right) 
                            \right]
                        \right]
        + \mathcal{R},
        \end{aligned}
    $}
    \end{adjustbox}
\end{equation}
where $p \left( y_{ij}|\mathbf{z}_i^{(u)}, \mathbf{z}_j^{(u)} \right)$ is defined in Equation (\ref{equ:y_real}) and $\mathcal{R}$ is defined in Equation~(\ref{equ:regularization}). 
We use the encoder $q(\cdot| \cdot)$, to disentangle the factors at test time. Since we only have access to the weak labels $\mathbf{Y}$ at training time, the encoder can only take $\mathbf{x}_{n}$ as an argument.
%Note that we let the encoder network $\mathbf{q}_{\boldsymbol{\phi}}(\mathbf{z}_n|\mathbf{x}_n)$ to be dependent on $\mathbf{x}_n$ only but independent of $\mathbf{Y}$ because when we test the model, we are usually provided with the instances $\mathbf{x}_n$ only without accessing $\mathbf{Y}$. 
%We summarize the encoder network in Figure \ref{fig:encoder}. 
%It is straightforward to solve the optimization problem in Equation \eqref{equ:objective} via stochastic gradient descent.
We use stochastic gradient descent (SGD) to optimize for $\boldsymbol{\theta}$ and $\boldsymbol{\phi}$.

\section{Related Work}
\label{sec:related_work}

There have been several unsupervised methods for learning disentangled representations with VAE, including $\beta$-VAE \cite{higgins2017beta}, factor VAE \cite{kim2018disentangling} , $\beta$-TCVAE \cite{chen2018isolating} and HCV \cite{lopez2018information}.  These methods achieve disentanglement by encouraging latent variables to be independent with each other. With these methods, the users can impact the disentanglement results only by tuning the hyperparameter. However, without explicit supervision, it is difficult to control the correspondence between a learned latent variable and a semantic meaning, and it is not guaranteed that the factor of interest can always be successfully disentangled \cite{locatello2018challenging}. In contrast, our proposed method utilizes the pairwise similarities as explicit supervision, which encourages the model to disentangle the factor of interest.

There have been attempts to improve disentanglement performance by introducing supervision.
\citet{narayanaswamy2017learning} and \citet{kulkarni2015deep} propose semi-supervised VAE methods that learn disentangled representation, by making use of partially observed class labels or real-value targets. \citet{bouchacourt2018multi} introduces supervision via grouping the samples. Our proposed method utilizes pairwise similarities.

Gaussian Process Prior VAE (GPPVAE) \cite{casale2018gaussian} assigns a Gaussian process prior to the latent variables. It makes use of the pairwise similarities between instances, by modeling the covariances between instances with a kernel function. GPPVAE does not focus on learning disentangled representation. Besides, GPPVAE requires the covariance matrix to be positive semi-definite, and the complete covariance matrix is observed without any missing values. In practice, a user might fail to provide labels satisfying these requirements. Our proposed method allows unobserved similarities and does not require the similarity matrix to be positive semi-definite. 

Dual Swap Disentangling (DSD) \cite{feng2018dual} and Generative Multi-view Model \cite{chen2018multiview}
are VAE and GAN models that make use of binary similarity labels, respectively. They both assume that the latent variables $\mathbf{z}$ can be separated into subspaces $\mathbf{z}^{(u)}$ and $\mathbf{z}^{(v)}$, which is similar to our proposed model. However, both methods assume that similar instances share similar $\mathbf{z}^{(u)}$, but do not force dissimilar instances to be encoded differently in $\mathbf{z}^{(u)}$. As shown in our experiments, DSD is likely to converge into a trivial solution that all instances share similar $\mathbf{z}^{(u)}$, despite the similarity labels. In contrast, our proposed model is able to make use of both binary and real-valued similarities and it avoids this trivial solution by utilizing both similarity and dissimilarity labels.

\section{Experiments}
In this section, we evaluate our method quantitatively and qualitatively. We perform experiments for both binary and real-value similarity values. Our method is compared against a few competing methods qualitatively in terms of recovering semantic factors for rotating object or identifying the labels on benchmark datasets, where we evaluate our approach quantitatively on the recovery of the ground-truth factors. Then we apply our method to analyze the real-world COPD dataset. Finally, we study the robustness of our method for the choice of hyperparameters, the proportion of the observed pairwise similarity and the noisiness of the observed similarities.

In the following, we first introduce datasets used for our experiments, followed by the discussion of the various quantitative metrics used in this paper. We then report the results of the experiments.

\subsection {Datasets and Competitive Methods}
\begin{table*}[t]
\caption{The Dataset}
\label{tbl:dataset}
\centering
\begin{adjustbox}{width=.75\textwidth, center}  
  \begin{tabular}{lcccc}
    \toprule
    Name   & Training instances & Held-out instances & Image size & The ground-truth factor \\
    \midrule
    MNIST & $60,000$ & $10,000$ &$28 \times 28 \times 1$ & discrete labels
    \\
    Fashion-MNIST  & $60,000$ & $10,000$  & $28 \times 28 \times 1$ & discrete labels
    \\
    Yale Faces & $1,903$ & $513$& $64 \times 64 \times 1$ & azimuth lighting 
    \\
    3D chairs & $69,131$ & $17,237$ & $64 \times 64 \times 3$ & azimuth rotations
    \\
    3D cars & $14,016$ & $3,552$ & $64 \times 64 \times 3$ & azimuth rotations
    \\
    \bottomrule
  \end{tabular}
\end{adjustbox}  
\end{table*}

We evaluate our methods on five datasets: MNIST~\cite{mnist}, Fashion-MNIST~\cite{xiao2017fashion}, Yale Faces~\cite{georghiades2001few}, 3D chairs~ \cite{aubry2014seeing} and 3D cars \cite{krause20133d}. The details of these datasets are summarized in Table~\ref{tbl:dataset}. For each dataset, we generate a subset of pairwise similarities based on one ground-truth factor of variations, as shown in the table. Unless specified otherwise, we let the number of observed pairwise labels be $0.01\%$ of the number of all possible pairs.
% We discuss how the number affects the performance of the model in Section \ref{sec:n_labels}.  
%For the MNIST and fashion-MNIST datasets, since discrete class labels are provided, we generate binary pairwise similarities based on whether one pair of instances belong to the same class or not. 
For the MNIST and fashion-MNIST datasets, we define $y_{ij} = \mathds{1} (t_i = t_j)$ where $t_{i}$ and $t_{j}$ are the ground-truth labels for the sample $i$ and $j$, and $\mathds{1}$ is the indicator function.
For Yale faces, 3D chairs and 3D cars, we use the Gaussian RBF kernel to define the similarities, \ie $y_{ij} = exp(- \delta (t_i, t_j) ^2  / \sigma^2 )$. Since the ground-truth factors in all three datasets involve azimuth angles, we use $\delta$ to denote the difference between two azimuth angles, {\it e.g.},~$\delta(350^{\circ}, 20^{\circ}) = 30^{\circ}$.

In addition to regular VAE~\cite{kingma2013auto}, we compare our proposed method with three disentanglement approaches based on VAE, including $\beta$-VAE~\cite{higgins2017beta}, factor VAE~\cite{kim2018disentangling} , $\beta$-TCVAE \cite{chen2018isolating}. 
%These variants are unsupervised methods that achieve disentangling by encouraging latent variables to be independent. 
As a supervised disentanglement method, we compare our approach with Dual Swap Disentangling (DSD)~\cite{feng2018dual}. The DSD is designed to analyze binary similarities and cannot be applied to real-valued similarities. To make all methods comparable, we use the same encoder and decoder architectures for all the methods, which include four convolutional layers and one fully connected layer.
To select the hyperparameters for our method, we use 5-fold cross validation on the training data. Since most of the competing methods are unsupervised, we choose the hyperparameters for them that achieves the best performance on the held-out data, which is advantageous for the competing methods resulting in an over-estimation of their performances. We define the metrics for the performance in the following section.

\subsection {Quantitative Comparison}
In this section, we perform two quantitative experiments. One is computing the Mutual Information Gap (MIG), which is a popular metric for evaluation of the disentanglement method, and the second experiment is a prediction task.

\noindent{\bf Mutual Information Gap (MIG)} \hspace{1mm}
We evaluate the disentanglement performance by computing the Mutual Information Gap (MIG) as introduced in~\cite{chen2018isolating}. Let $t$ represent the ground-truth factor and $\mathcal{I}(\cdot, \cdot)$ represent the mutual information between two random variables (with 1 or more dimensions). In our model, since we assume $\mathbf{z}^{(u)}$ is relevant to $t$, we expect $\mathcal{I}(\mathbf{z}^{(u)}; t)$ to be large; while 
$\mathcal{I}(\mathbf{z}^{(v)}_{\cdot d}; t)$ to be small for each dimension 
$d \in \{1 \ldots d^{(v)}\}$. 
Therefore, we can measure the disentanglement by computing the mutual information gap, defined as
\begin{equation}
    \small
    \label{equ:MIG}
    \frac{1}{ \mathcal{H}(t) } \left( \mathcal{I} (\mathbf{z}^{(u)}; t) - \max_{d \in \{1 \ldots d^{(v)}\}} \mathcal{I} (\mathbf{z}^{(v)}_{\cdot d}; t) \right),
\end{equation}
where $\mathcal{H}(\cdot) $ represents the entropy of a random variable. The values of  $\mathcal{I}(\cdot, \cdot)$ and $\mathcal{H}(\cdot) $ can be empirically estimated as explained in~\cite{chen2018isolating}. For each dataset, the dimensionality of $\mathbf{z}^{(u)}$, denoted by $d^{(u)}$, is shown in the Table~\ref{tbl:MIG}.
Our method directly produces the $\mathbf{z}^{(u)}$ and $\mathbf{z}^{(v)}$ terms that can be plugged into Equation~(\ref{equ:MIG}). Since the competing methods are unsupervised, the choice of the indices for $\mathbf{z}^{(u)}$ and $\mathbf{z}^{(v)}$ is not clear. For those methods, we first rank all latent variables based on the mutual information with respect to the ground-truth. Then, we pick  the top $d^{(u)}$ random variables to form $\mathbf{z}^{(u)}$ and the remaining latent variables are assigned to $\mathbf{z}^{(v)}$. The MIG values are estimated on the held-out data.

The values in Table~\ref{tbl:MIG} report the MIG for various methods. Our proposed method achieves substantially higher MIG values than other approaches. It outperforms the second-best methods by more than $40\%$ in all five datasets. The results illustrate the importance of introducing supervision in disentanglement tasks. Although DSD is a supervised method that is formulated to incorporate binary pairwise similarities, it fails to disentangle the ground-truth factor. We speculate that the failure is due to convergence to a trivial solution, as mentioned in the Related Work Section. 

\noindent{\bf Prediction Task } \hspace{1mm}
We use $\mathbf{z}^{(u)}$ as an input to a regression or classification method to predict the ground truth. We use the $5$ Nearest Neighbour ($5$-NN) method for both classification and regression. Table~\ref{tbl:downstream} reports the outcome for different datasets, measured by  Cohen's kappa ($\kappa$)  and $R^2$ with respect to the ground-truth. We measure Cohen's kappa rather than classification accuracy because it corrects for the possibility of the agreement occurring by chance. For both measurements, a higher value  indicates a better performance. We observe that our proposed method outperforms the competing methods in all tasks. This implies that instances with similar ground-truth factors are located near each other in the latent space $\mathbf{z}^{(u)}$.

\begin{table*}[t]
\centering
\caption{ MIG metrics on the held-out data}
\begin{adjustbox}{width=.68\textwidth, center}  
    \begin{tabular}{l m{1.5cm} c c c c c c	}
    \toprule
    Dataset & $d^{(u)}$ & Proposed & VAE& $\beta$-VAE & Factor-VAE & TCVAE & DSD
    \\
    \midrule
    MNIST 
    & $2$ & $ \mathbf{ 0.68}$ & $0.01$ &$0.03$ & $0.33$ & $0.04$ & $0.01$
    \\
    Fashion-MNIST 
    & $2$ & $ \mathbf{0.52}$ & $0.11$ & $0.28$ & $ 0.36 $ & $0.19$ & $0.01$
    \\
    Yale Faces 
    & $1$ & $\mathbf{0.42}$ & $0.02$ &$0.07$ & $0.06$ & $0.29$ & N/A \footnotemark[1]
    \\
    3D chairs 
    & $2$ & $\mathbf{0.37}$ & $0.02$ &$0.15$ & $0.11$ & $0.08$ & N/A \footnotemark[1]
    \\
    3D cars 
    & $2$ & $\mathbf{0.41}$ & $0.02$ &$0.22$ & $0.15$ & $0.16$ & N/A \footnotemark[1]
    \\
    \bottomrule
    \multicolumn{8}{l}{\footnotemark[1] DSD is designed for analyzing binary similarities, and cannot analyze real-valued similarities.}
    \label{tbl:MIG}
    \end{tabular}
\end{adjustbox}
\end{table*}

\begin{table*}[t]
    \caption{Prediction Performance }
    \label{tbl:downstream}
    \centering
\begin{adjustbox}{width=.68\textwidth, center}  
    \begin{tabular}{c l c c c c c c	}
    \toprule
    \small
    & Dataset & Proposed & VAE& $\beta$-VAE & Factor-VAE & TCVAE & DSD
    \\
    \midrule
    \multirow{2}{*}{$\kappa$} &
    MNIST 
    & $ \mathbf{ .969 }$ 
    & $.494$ &$.326$ & $.704$ & $.260$ & $ .030 $
    \\
    & Fashion-MNIST 
    & $ \mathbf{.857}$ & $.389$ & $.460$ & $ .613 $ & $.482$ & $.003$
    \\
    \midrule
    \multirow{3}{*}{$R^2$} &
    Yale Faces 
    & $\mathbf{.968}$ & $.397$ &$.760$ & $.699$ & $.692$ & N/A \footnotemark[1]
    \\
    & 3D chairs 
    & $\mathbf{.912}$ & $.155$ &$.357$ & $.224$ & $.196$ & N/A \footnotemark[1]
    \\
    & 3D cars 
    & $\mathbf{.584}$ & $.391$ & $.418$ & $.177$ & $.110$ & N/A \footnotemark[1]
    \\
    \bottomrule
    \multicolumn{8}{l}{\footnotemark[1] DSD is designed for analyzing binary similarities, and cannot analyze real-valued similarities.}    
    \end{tabular}
\end{adjustbox}    
\end{table*}

\begin{figure*}[t!]
\includegraphics[width=1.0\textwidth]{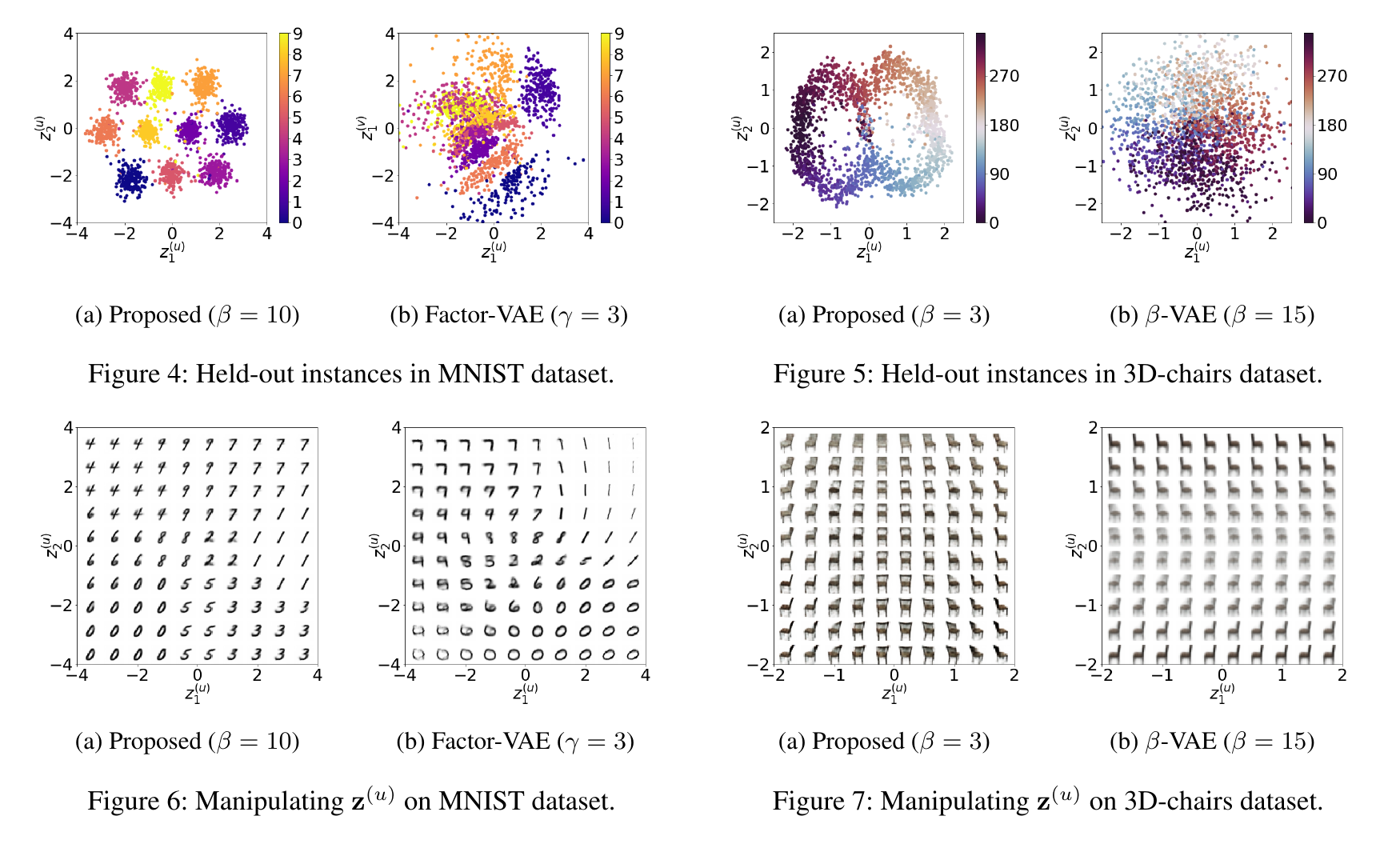}
\end{figure*}

\subsection {Qualitative Comparison}
In this subsection, we illustrate the disentanglement performance of the proposed method via qualitative comparison. We use the results on the MNIST and 3D-chairs datasets as examples (for more experimental results, see the supplementary materials \textsuperscript{\rm 1} \footnote{ \textsuperscript{\rm 1} Supplementary materials are available at \url{https://arxiv.org/abs/1906.01044} }).

\noindent{\bf MNIST} \hspace{1mm}
Figure 4(a) demonstrates $\mathbf{z}^{(u)}$ of the held-out instances from the MNIST dataset. Different colors represent different class labels. Figure 4(b) shows a similar concept for the competing method that achieves the highest MIG value in Table~\ref{tbl:MIG}.
We observe that the proposed model is able to learn $\mathbf{z}^{(u)}$ such that it explains the ground-truth factor (\ie the digit class). 
All ten classes are well separated in the latent space with distinct centers, and instances from the same class are located close to each other. 
As shown in Figure 4(b), the factor-VAE is also able to learn a disentangled representation. However, regions of the instances of digit $4$ and $9$ are overlapping in the latent space. 
%The disentangling performance of factor-VAE is not as good as the proposed method.

To illustrate the performance of the generative model, we plot some images generated by the proposed and the competing method in Figure~6(a) and~6(b), respectively. We first randomly sample an image from the held-out data and encode it into $\mathbf{z} = [ \mathbf{z}^{(u)}, \mathbf{z}^{(v)}]$. Then, we keep $\mathbf{z}^{(v)}$ constant and manipulate $\mathbf{z}^{(u)}$. Using the new code, we generate new images that are displayed at their corresponding locations.
In Figure~6(a), we find that the writing styles of ten digits are similar. This implies that $\mathbf{z}^{(u)}$ only contains the information about the ground-truth factor and not the other factors of variation. In contrast, we observe changes in writing styles in Figure~6(b). The figure shows that the reconstructed digits have different thicknesses, angles, widths.

\noindent{\bf 3D-chairs} \hspace{1mm} 
We repeat the same plotting process for the 3D-chairs dataset. The results are shown in Figures 5 and 7.
Since the ground truth ( {\it i.e.,} azimuth ) is a cyclic value, the ideal shape of the latent variable should look like a ring, which is approximately captured by our method in Figure~5(a).
For some images, it is more challenging to determine which direction the chair faces (some chairs are almost centrosymmetric). These images are encoded into the regions close to the origin. Without proper supervision, $\beta$-VAE is not able to fully recover the complex underlying structure of the ground-truth factor, as shown in Figure~5(b).

We manipulate $\mathbf{z}^{(u)}$ and generate the images in Figure~7. As shown in~7(a), we observe the images of chairs facing various directions, located at the ring displayed in Figure~5(a). In Figure~7(b), we observe that $\beta$-VAE can reconstruct the chair images facing left and right, but other reconstructed images are blurry.

\subsection{COPD dataset}
\label{sec:COPD}

\begin{table}[t]
\centering
    \captionof{table}{ Prediction Performance in the COPD dataset}
    \begin{adjustbox}{width=\columnwidth, center}  
        \begin{tabular}{l l c c c c c}
            \toprule
            & & Proposed & VAE & $\beta$-VAE & Factor-VAE & TCVAE \\
            \midrule
            \multirow{3}{*}{$R^2$} & 
            $\text{FEV}_1$pp
            & \bf{.431} & .002 & .013 & .010 & .040
            %\\ 
            %& $\text{FEV}_1 / \text{FVC}$ 
            %& \bf .578 & .060 & .137 & .083 & .086
            \\
            & Emphesyma\% 
            & \bf .441 & .027 & .252 & .191 & .081
            \\
            & GasTrap\%
            & \bf .522 & .104 & .279 & .067 & .110
            \\
            \toprule
            $\boldsymbol{\kappa}$ 
            & GOLD 
            & \bf .236 & .023 & .089 & .061 & .088
            \\
            \bottomrule
        \end{tabular}
        \label{tbl:prediction}
    \end{adjustbox}
\end{table}

A real-world application of the proposed model is to analyze the COPD dataset. The purpose of this application is to identify factors in the Computer Tomography (CT) images of the chest that are related to disease severity. We applied our method on a large-scale dataset (over 9K patients), where all patients have CT images as well as spirometric measurements. We use the spirometric measures to construct pairwise real-value similarities using Radial Basis Function. 

In the COPD dataset, a ground-truth measure for disease severity is not available. Therefore, we use $\mathbf{z}^{(u)} \in \mathbb{R}$ learned by our model to predict several clinical measurements of disease severity from different aspects, via a 5-nearest neighbor regression and classification. The clinical measurements include (1)~\texttt{ $\text{FEV}_1$pp} measuring how quickly one can exhale, (2)~\texttt{Emphesyma\%} measuring the percentage of destructed lung tissue, (3)~\texttt{GasTrap\%} indicating amount of gas trapped in lung, and (4)~\texttt{GOLD} score which is a six-categorical value indicating the severity of airflow limitation. In Table~\ref{tbl:prediction}, we report $R^2$ for the first three measurements and Cohen's kappa coefficient ($\kappa$) for the last measurement. The results suggest that our method is better than the unsupervised approach in disentangling the disease factor, as it outperforms them in predicting various measures of disease severity.

\subsection{Choice of Hyperparameters}
\label{sec:beta}
To illustrate how the hyperparameter $\beta$ affects the performance of our proposed method, we first plot generated images with an improperly chosen $\beta$ in Figure 8. In this figure, we find all ten digits. However, unlike the results shown in Figure 6(a), the writing styles (thicknesses, angles, widths, sizes, {\it etc.}) of the generated digits change significantly. This implies a failure of disentanglement, because $\mathbf{z}^{(u)}$ explains some factors of variations other than the one of interest ({\it i.e.}, digit class).  

To find a proper $\beta$ for each dataset, we vary $\beta$ and conduct $5$-fold cross validation with the training instances. We plot the mean log-likelihood ( $\log p_{\boldsymbol{\theta}} (\mathbf{X}, \mathbf{Y} | \mathbf{Z})$ ) of five validations sets in Figure~9. We observe that a maximum log likelihood is achieved with choices of $\beta$ between $2$ to $10$, but the optimal $\beta$ differs across datasets. We choose $\beta$ that maximizes the log-likelihood for each dataset. 

We illustrate how the hyperparameters $\eta_1$ and $\eta_2$ affect the disentanglement performance in Figure~10.
In Figure~10(a), we fix $\eta_2 = 2$ and vary $\eta_1$; while in Figure~10(b), we fix $\eta_1 = 1e3$ and vary $\eta_2$. Because the log-likelihood is a function of $\eta_1$ and $\eta_2$, we report the MIG metrics for the held-out data, instead. We observe that when $\eta_1 \ge 1e3$ and $\eta_2 \ge 1.$, these hyperparameters have limited effects on the MIG metrics. In all other experiments, we choose $\eta_1 = 1e3$ and $\eta_2 = 2$.

\begin{figure}[t]
    \centering
    \includegraphics[width = 1.\columnwidth]{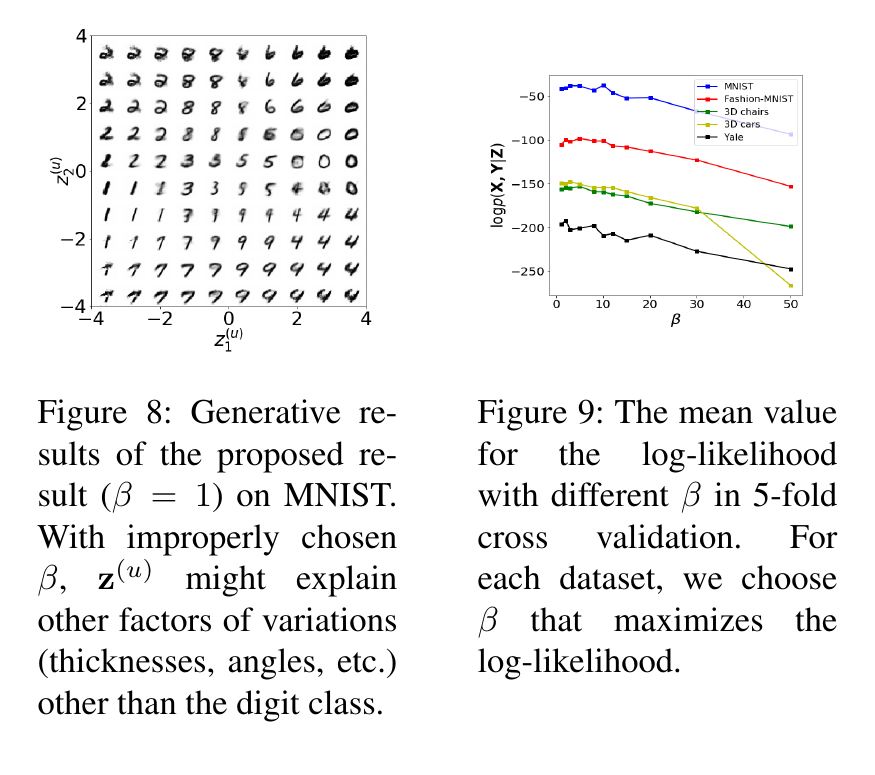}
\end{figure}

\begin{figure}[t]
    \centering
    \includegraphics[width = 1.\columnwidth]{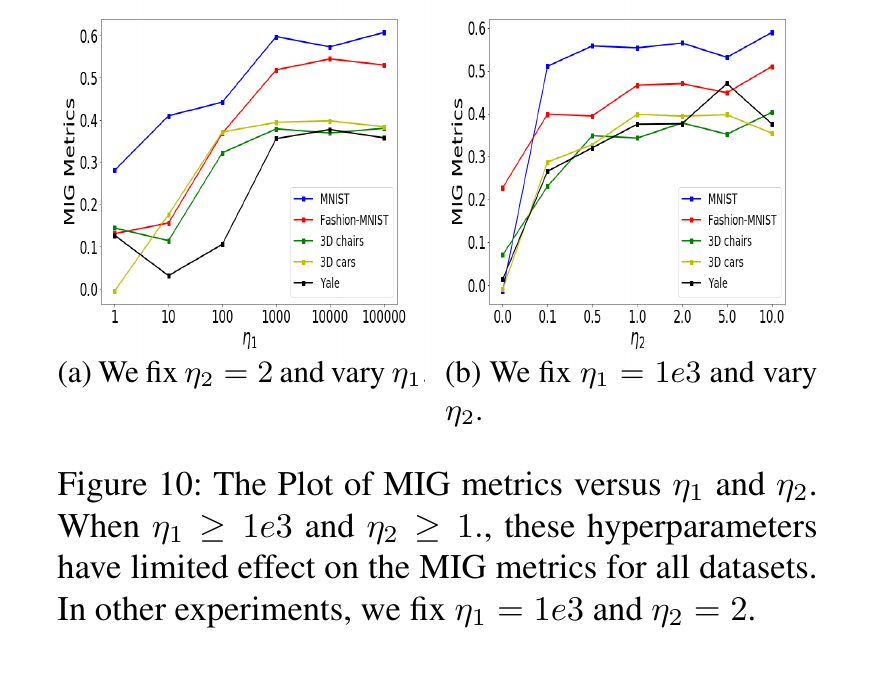}
\end{figure}

\begin{figure}[t]
    \centering
    \includegraphics[width = 1.\columnwidth]{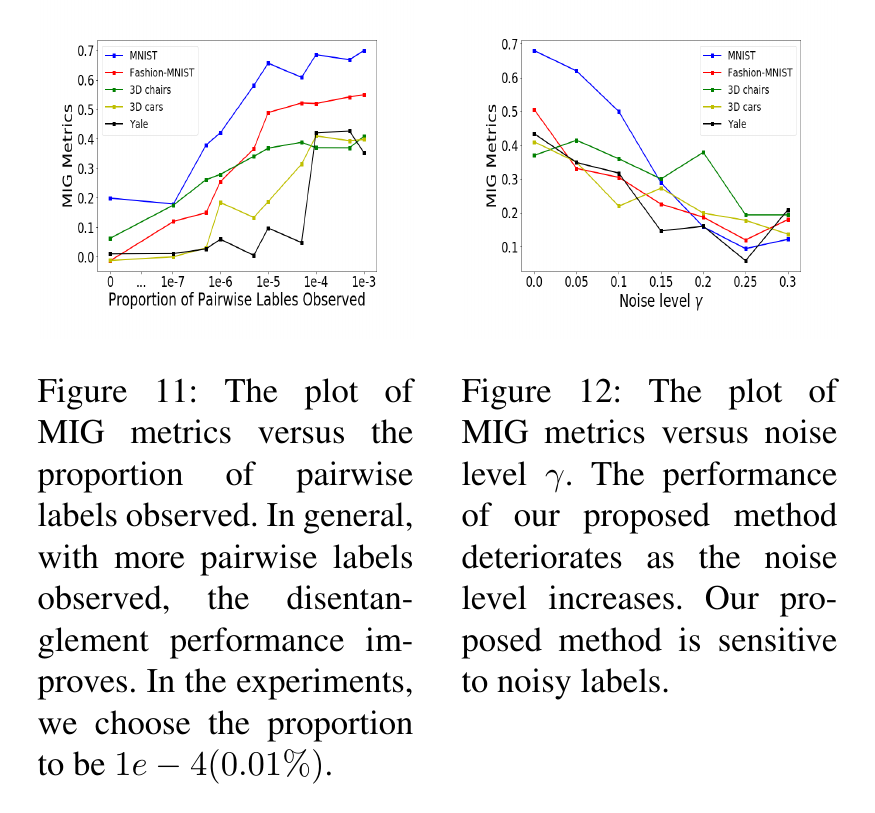}
\end{figure}

\subsection{Number of Pairwise Labels}
\label{sec:n_labels}
We investigate how the number of pairwise labels affects the performance of our proposed model. In Figure~11, we plot the MIG metrics for the held-out data versus the proportion of observed pairwise labels in training. We observe that in general, with more pairwise labels provided, the disentanglement performance improves. However, as the proportion approaches $1e-4$, the rate of improvement tapers. In all other experiments, we fix the proportion to be $1e-4$. 

\subsection{Noisy Similarity Labels}
\label{sec:noisy_labels}
In all previous experiments, we do not introduce noise to the pairwise similarity labels. In this section, we introduce noise controlled by the noise level $\gamma$. For binary labels, we flip the labels with probability $\gamma$. For real-valued similarities, we let $\gamma$ be the variance of the Gaussian noise, {\it i.e.}, we add Gaussian noise $\epsilon \sim \mathcal{N}(0, \gamma )$ and clip the results. We observe Figure~12 that the performance of our proposed method deteriorates as the noise level increases. Our proposed method is sensitive to noisy labels. By comparing the results to values in Table \ref{tbl:MIG}, we conclude that when $\gamma \le 0.1$, our proposed method gives better or comparable MIG metrics than the competing methods.

\section{Conclusion}

In this paper, we investigate the disentanglement learning problem, assuming a user introduces weak supervision by providing similarities between instances based on a factor to be disentangled.  The similarity is provided as either a discrete (yes/no) or real-valued label between $0$ and $1$, where a larger value indicates a stronger similarity. We propose a new formulation for weakly supervised disentanglement of latent variables within the Variational Auto-Encoder (VAE) framework. Experimental results on both benchmark and real-world datasets demonstrate that utilizing weak supervision improves the performance of VAE in disentanglement learning tasks.

\section*{Acknowledgments}

This work was partially supported by NIH Award Number 1R01HL141813-01, NSF 1839332 Tripod+X, and SAP SE. We gratefully acknowledge the support of NVIDIA Corporation with the donation of the Titan X Pascal GPU used for this research. We were also grateful for the computational resources provided by Pittsburgh SuperComputing grant number TG-ASC170024.

\bibliographystyle{aaai}
\bibliography{AAAI-ChenJ.1491}



\section*{Supplementary material}

# Supplementary Material

## 1 The Plots in the Latent Space

**Fashion-MNIST dataset** and **3D cars** We include the plots $\mathbf{z}^{(u)}$ for the held-out instances and the generative instances by manipulating $\mathbf{z}^{(u)}$, for fashion-MNIST and 3D cars-datasets. We observe that although it is noisy, most of the instances are well separation according to the ground-truth factors. This implies that $\mathbf{z}^{(u)}$ can explain the ground-truth factors.

It is not visually clear whether $\mathbf{z}^{(u)}$ explains other factors of variations else than the ground-truth factors, as shown in Figure 1. In Figure 2b, we observe that the shape and color of the cars do not change in the rotations. In this example, $\mathbf{z}^{(u)}$ does not seem to explain other factors other than the ground-truth factor.

**Yale Faces dataset** Since we choose $d^{(u)} = 1$ for the Yale dataset, in Figure 4c, we plot $\mathbf{z}^{(u)}$ against one dimension in $\mathbf{z}^{(v)}$. we observe that most instances with different ground-truth values can be separated according to $\mathbf{z}^{(u)}$. We manipulate $\mathbf{z}^{(u)}$ and one dimension $\mathbf{z}^{(v)}$ and plot the generated images in Figure 4e. We observe that the identify of the face changes when we change $\mathbf{z}^{(v)}$. However, when we change $\mathbf{z}^{(u)}$, the azimuth lighting changes, but the identity does not change. This suggests good disentangling results for this dataset.

## 2 Reconstructed and Generated Images

In Figure 4, we plot some reconstructed and generated images. The first row of each of the subplots are the original images in the held-out data. The second row shows the reconstructed image. On the right-hand side of these subplots, we show the reconstructed images by manipulating the latent variables $\mathbf{z}^{(u)}$ but keep $\mathbf{z}^{(v)}$ constants. We observe that our proposed model does not always reconstruct the images perfectly, probably because the models have never observed the held-out instances. However, for the generated images on the right-hand side, we observe the factor of interest changes, but other factors of variations do not change as much. In particular, the factor of interest is similar for those images with the same $\mathbf{z}^{(u)}$ values. This implies excellent disentangling performance.

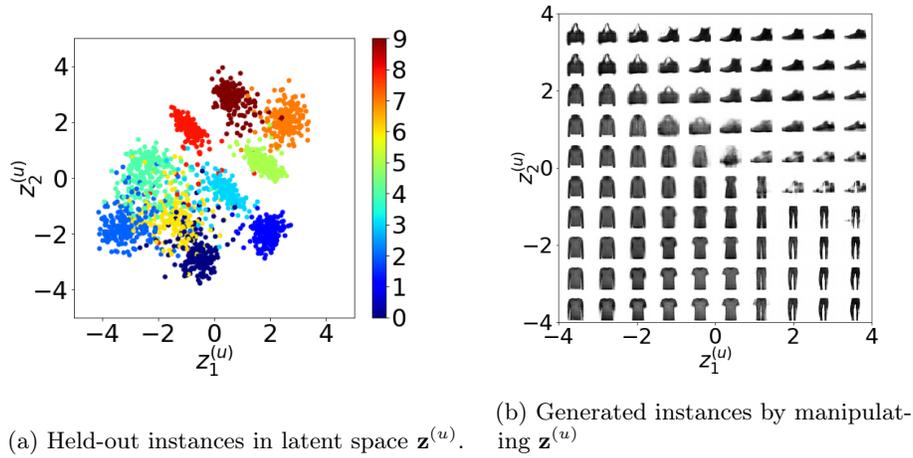

(a) Held-out instances in latent space $\mathbf{z}^{(u)}$.

(b) Generated instances by manipulating $\mathbf{z}^{(u)}$

Figure 1: Fashion-MNIST dataset

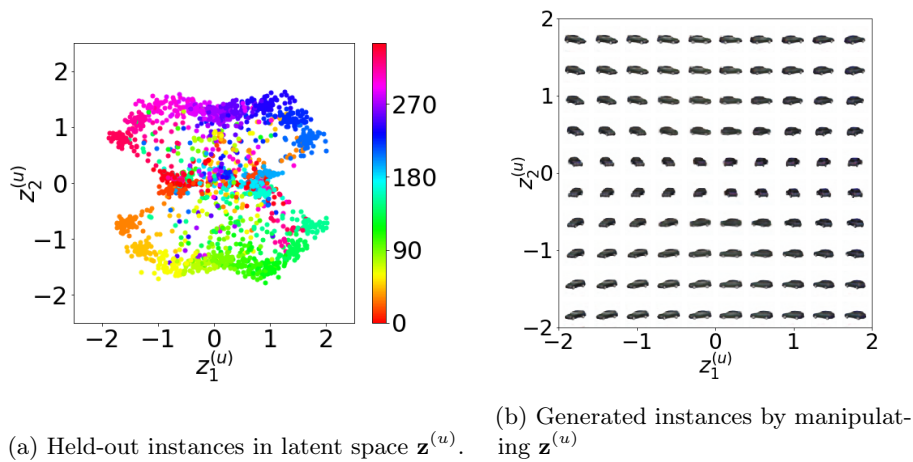

(a) Held-out instances in latent space $\mathbf{z}^{(u)}$.

(b) Generated instances by manipulating $\mathbf{z}^{(u)}$

Figure 2: 3D cars dataset

2

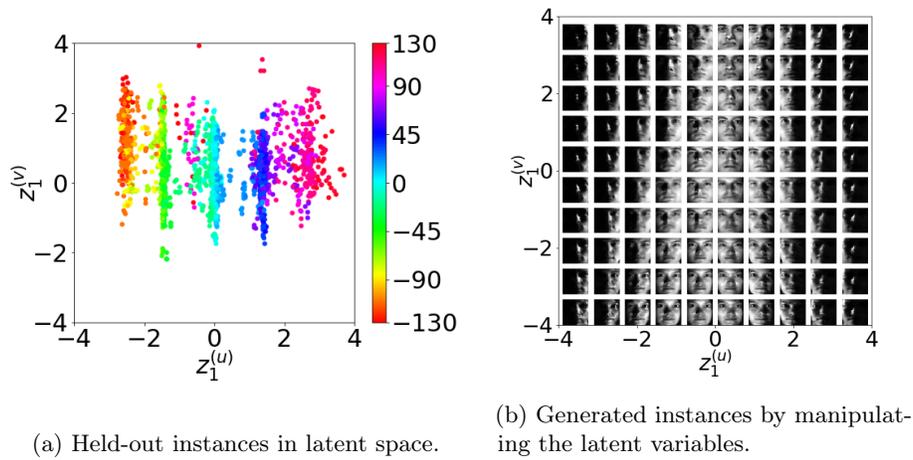

(a) Held-out instances in latent space.

(b) Generated instances by manipulating the latent variables.

Figure 3: Yale Faces dataset

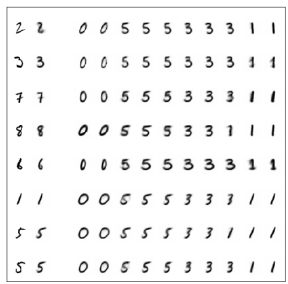
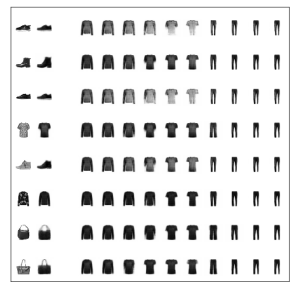

(a) MNIST dataset

(b) Fashion-MNIST dataset

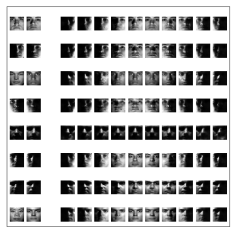
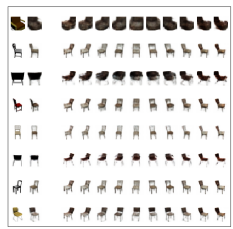
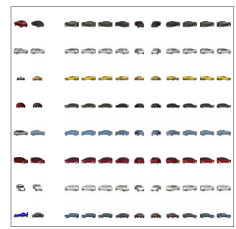

(c) Yale faces Dataset  (d) 3D chairs dataset  (e) 3D cars dataset

Figure 4: Reconstructed and Generated Images

3


\end{document}